\documentclass[10pt]{article}

\usepackage[dvips]{graphicx}
\usepackage{amssymb,amsfonts,amsmath,amsthm}
\usepackage{psfrag}
\usepackage[numbers]{natbib}

\newcommand{\fs}[1]{\mathcal{#1}}

\newcommand{\order}{\mathcal{O}}
\newcommand{\define}{:=}
\newcommand{\comment}[1]{}

\newcommand{\expect}{\mathbf{E}}

\newcommand{\g}[1]{\underline{#1}}
\newcommand{\finint}{\fs{Z}^\ast}   
\newcommand{\infint}{\fs{Z}^\infty} 
\newcommand{\prob}{\mathbf{P\!r}}      

\theoremstyle{plain}

\theoremstyle{definition}

\begin{document}


\title{Generalized Thompson sampling for sequential decision-making and causal inference}
\author{Pedro A. Ortega and Daniel A. Braun}

\maketitle

\begin{abstract}
Recently, it has been shown how sampling actions from the predictive distribution over the
optimal action---sometimes called \emph{Thompson sampling}---can be applied to solve sequential adaptive control problems, when the optimal policy is known for each possible environment. The predictive distribution can then be constructed by a Bayesian superposition of the optimal
policies weighted by their posterior probability that is updated by Bayesian inference and causal calculus. Here we discuss three important features of this approach. First, we discuss in how far such Thompson sampling can be regarded as a natural consequence of the Bayesian modeling of policy uncertainty.
Second, we show how Thompson sampling can be used to study interactions between multiple adaptive agents, thus, opening up an avenue of game-theoretic analysis. Third, we show how Thompson sampling can be applied to infer causal relationships when interacting with an environment in a sequential fashion. In summary, our results suggest that Thompson sampling might not merely be a useful heuristic, but a principled method to address problems of adaptive sequential decision-making and causal inference.
\end{abstract}


\section{Introduction}

In a research paper from $1933$, Thompson studied the problem of finding
out which one of two drugs was better when testing them on a patient population under the constraint that as few people as possible should be subjected to the inferior drug during the course of testing \cite{Thompson1933}. Given a current probability estimate $P$ of one treatment being better than the other, he suggested that it might be a good idea to adjust the proportions of future test subjects that take the two drugs to the respective probabilities $P$ and $1-P$. This way one would not run into the danger of \emph{permanently} cutting off all future test subjects from
a potentially superior treatment that so far seems inferior due to statistical fluctuations, while only \emph{temporarily} risking exposure to a potentially inferior drug for a smaller proportion of the population. Randomizing actions based on the probability that this action is believed to be optimal when faced with an unknown environment is now sometimes called \emph{Thompson sampling}.

Thompson sampling is a form of \emph{probability matching}. Probability matching has been extensively studied in both humans and animals when they make predictions in stochastic environments \cite{Myers1976,Vulkan2000}. Rather than consistently predicting the most likely outcome, experimental subjects tend to randomize their predictions based on the probabilities with which the respective events occur. When knowing the probabilities, this is clearly a suboptimal strategy. However, in the case of Thompson sampling it is important to note that the probabilities are not known. Nevertheless, one might argue that Thompson sampling is a suboptimal strategy, as Thompson's problem can be thought of as a \emph{bandit problem} \cite{Sutton1998}, which is solved optimally by Gittin's indices in the case of known prior probabilities and discounted rewards \cite{Gittins1979}.
Most studies have therefore examined Thompson sampling as a heuristic in the context of
bandit problems \cite{Granmo2008,Asmuth2009,Granmo2010,Graepel2010,Scott2010,May2011,Chapelle2011,Agrawal2011,Granmo2012,May2012,russo2013}.

Recently, it was shown, however, that Thompson sampling can also be applied to solve a more general class of sequential adaptive control problems, provided that both an optimal policy and a predictive model is known for each possible environment \cite{OrtegaBraun2010e}. When an environment is drawn randomly from the set of possible environments, the optimal policy
can then be inferred on the fly by an adaptation process that is driven by actions sampled from the predictive distribution over the optimal policies.
Here we study three characteristic features of such generalized Thompson sampling. First, we discuss in how far Thompson sampling can be regarded as a natural consequence of a Bayesian treatment of policy uncertainty. Second, we study convergence behavior when two adaptive Thompson sampling agents are coupled in a sequential fashion. Third, we show how this approach can be extended naturally to address the problem of causal induction when interacting with an unknown environment.

The paper is structured as follows.
In Section~2 we clarify the problem statement and recapitulate the main result of \cite{OrtegaBraun2010e}. In Section~3 we analyze the uncertainty faced by agents that are unable to compute the single best policy. In Section~4 we study interactions that arise when coupling two adaptive agents that employ Thompson sampling to determine their actions. In Section~5 we investigate how this approach can be applied to adaptive agents that need to discover the causal structure of their environment.
Finally, we discuss in what sense Thompson sampling might provide a principled solution to adaptive decision-making problems.

\section{Problem Statement}
\label{sec:problem}
\subsection{Preliminaries}

We restrict the exposition to the case of discrete time with discrete
stochastic observations and actions. Let $\fs{O}$ and $\fs{A}$ be two
finite sets, the first being the \emph{set of observations} and the second
being the \emph{set of actions}. We use $a_{\leq t} \define a_1 a_2 \ldots
a_t$ and $a_{< t} \define a_1 a_2 \ldots
a_{t-1}$  to simplify the notation of
strings. We define the \emph{set of interactions} as $\fs{Z} \define \fs{A}
\times \fs{O}$. The
set of interaction strings of length $t \geq 0$ is denoted by $\fs{Z}^t$.
The set of all finite interaction strings is $\finint
\define \bigcup_{t \geq 0} \fs{Z}^t$, the set of infinite strings is $\infint \define
\{ w: w = a_1 o_1 a_2 o_2 \ldots \}$. The interaction string of length 0 is
denoted by $\epsilon$.

Agents and environments are formalized as I/O systems. An \emph{I/O system}
$\prob$ is a probability distribution over interaction sequences $\infint$.
$\prob$ is uniquely determined by the conditional probabilities
\[
    \prob(a_t|a_{<t},o_{<t}), \quad \prob(o_t|a_{\leq t},o_{<t})
\]
for each $a_1 o_1 \ldots a_{t-1} o_{t-1} a_t \in \finint$. An \emph{interaction system}
$(P, Q)$ is a coupling of two I/O systems, where $P$ is an agent and $Q$ is an environment.
Because the agent and the environment mutually influence each other, their actions and observations are conditioned by the previous interactions. Accordingly, the probability of an interaction string $a_1 o_1 \ldots a_T o_T$ is given by
\begin{equation}\label{eq:prob-interactions}
    \prod_{t=1}^T P(a_t|a_{<t},o_{<t})
        Q(o_t|a_{\leq t},o_{<t}).
\end{equation}

From the point of view of the agent $P$, the distribution
$P(a_t|a_{<t},o_{<t})$ is a \emph{policy} and captures the probability of producing action $a_t \in
\fs{A}$ given history $a_1 o_1 \ldots a_{t-1} o_{t-1}$. The distribution $P(o_t|a_{\leq t},o_{<t})$ is the agent's \emph{predictive model} of the environment, as it predicts the probability of the observation $o_t \in \fs{O}$ given history $a_1 o_1 \ldots a_{t-1} o_{t-1} a_t$.
For the agent $P$, the sequence $o_1 o_2 \ldots$ provides its
input stream and the sequence $a_1 a_2 \ldots$ is its output stream. In the case of the
environment $Q$ the roles are reversed, that is the sequence $o_1 o_2 \ldots$ is its output
stream and the sequence $a_1 a_2 \ldots$ provides its input stream.
The quintessential goal is to choose the agent's policy such that the resulting distribution over the interaction sequences \eqref{eq:prob-interactions} is \emph{desirable}.

\subsection{Policy Construction: Known Environment}

If $Q$ is known, then $P$ can be equipped with a model
that can perfectly predict its environment, that is
$P(o_t|a_{\leq t},o_{<t}) = Q(o_t|a_{\leq t},o_{<t})$ for all $a_1 o_1 \ldots a_t \in \finint$.
Moreover, a custom-made policy can be designed for $P$ that produces desirable interaction sequences. Desirability is typically formalized by the economic theory of subjective expected utility (SEU) \cite{Neumann1944,Savage1954}, which stipulates
that a decision maker's preferences over lotteries \emph{can be thought of} as maximizing a SEU of the outcome. In the policy construction setting, this translates into the designer having a real-valued \emph{utility function} giving rise to utilities $U(a_{1:T}, o_{1:T})$ for each realization, and the \emph{predictive model} $P(o_t|a_{\leq t},o_{<t})$. The utility function quantifies the subjective desirability of a particular interaction string and the probabilities represent the subjective model of the environment. The \emph{maximum expected utility principle} then states that the designer has to choose the policy such that it maximizes the expected utility
\[
    \expect_P[U] :=
    \sum_{a_{\leq T}, o_{\leq T}}
        P(a_{\leq T}, o_{\leq T}) U(a_{\leq T},o_{\leq T})
\]
where the probabilities $P(a_{\leq T}, o_{\leq T})$ are the policy-prediction products
\begin{gather}
    \nonumber
    \prod_{t=1}^T  P(a_t|a_{<t},o_{<t})
        P(o_t|a_{\leq t},o_{<t}) \\
    = \biggl\{
        \underbrace{
            \prod_{t=1}^T P(a_t|a_{<t},o_{<t})
        }_\text{policy}
    \biggr\}
    \biggl\{
        \underbrace{
            \prod_{t=1}^T P(o_t|a_{\leq t},o_{<t})
        }_\text{prediction}
    \biggr\}.
\end{gather}
The optimal policy is often computed by restating the problem recursively and then using dynamic programming to solve the Belmann optimality equations \citep{Bellman1957}. The policy and the prediction model are both subjective in the sense that they are unilaterally chosen by the designer. A policy choice explainable by this scheme is defined to be a \emph{rational} choice. Choices that do not strictly obey the maximum SEU principle are \emph{irrational}, or at best \emph{bounded rational} \citep{Simon1957}.

\subsection{Policy Construction: Unknown Environment}
\label{sec:SEUformulation}
In general, the prediction model will not be equal to the generative law of the environment, that is,
\[
    P(o_t|a_{\leq t},o_{<t}) \neq Q(o_t|a_{\leq t},o_{<t}).
\]
and consequently the true expected utility is in general not equal to the SEU:
\begin{equation}
    \expect_\mathrm{Pr}[U]
    \neq \expect_P[U].
\end{equation}
One of the most interesting cases where the prediction model and the true generative law of the environment do not match is when the designer is uncertain about the latter. Formally, the designer expresses his uncertainty by introducing a random variable $\theta$ that indexes the class of potential environments $Q_\theta$. More specifically, he has a class of prediction models and policies
\begin{equation}\label{eq:prediction-models}
    B(o_t|\theta, a_{\leq t}, o_{< t}), \quad B(a_t|\theta, a_{< t}, o_{< t}),
\end{equation}
such that for every possible environment indexed by $\theta$ there is a perfectly fitting predictor $B(o_t|\theta, a_{\leq t}, o_{< t}) = Q_\theta(o_t|a_{\leq t}, o_{< t})$ and a desirable custom-built policy $B(a_t|\theta, a_{< t}, o_{< t})$. Moreover, the designer believes that
$Q_\theta$ is drawn with probability $B(\theta)$ from a set $\Theta$
of possible environments before the interaction starts, where $\Theta$ is assumed to be discrete for simplicity.
\subsubsection{Decision-theoretic Problem Formulation}

In order to stay within the framework of subjective expected utility one has to reduce the problem of the unknown environment to a problem with known environment. Such a ``known'' environment can be created from a set of possible environments as a new ``super-environment'' by marginalizing over the parameter of the possible environments, thus, obtaining the \emph{Bayesian mixture distribution} \citep{Hutter2004}
\begin{align*}
    B(o_{\leq T}|a_{\leq T})
    &= \sum_\theta B(\theta) B(o_{\leq T}|\theta, a_{\leq T})
    \\&= \sum_\theta B(\theta) \prod_{t=1}^T
        B(o_t|\theta, a_{\leq t}, o_{<t}).
\end{align*}
The adaptive control problem is then solved
by equating the prediction model $P$ with the Bayesian predictive distribution over the observations,
\[
    P(o_t|a_{\leq t}, o_{<t}) := B(o_t|a_{\leq t}, o_{<t}),
\]
and then choosing the policy that maximizes the SEU as in the case of a known environment. This procedure effectively enlarges the space of possible environments to the convex hull $H(\Theta)$ spanned by the prediction models in $\Theta$, i.e.\ an ``environment'' is any convex combination of distributions indexed by $\theta$. Each $\theta \in H(\Theta)$ thus corresponds to a Bayesian mixture over the environments in $\Theta$, but with a different prior. If the true generative law coincides with one of the models, i.e.\ there is a $\theta^\ast \in \Theta$ such that
\begin{equation}\label{eq:true-prediction-model}
    Q(o_t|a_{\leq t}, o_{<t}) = B(o_t|\theta^\ast, a_{\leq t}, o_{<t}),
\end{equation}
then the predictive model will converge to the true generative law with $\mathrm{Pr}$-probability one, that is
\[
    P(o_t|a_{\leq t}, o_{<t})
    \rightarrow Q(o_t|\theta^\ast, a_{\leq t}, o_{<t})
\]
as $t \rightarrow \infty$. Since the policy choice is optimal, this scheme directly bypasses the \emph{exploration-exploitation dilemma} \citep{Duff2002}.
The important point, however, is
that in order to express the uncertainty over the environment's probability law, the designer had to introduce a belief model that \emph{compiles} into an actual prediction model. Thus, the policy construction in the SEU problem statement for unknown environments proceeds in two steps: first, a Bayesian mixture environment is created, and second a utility-maximizing policy is found on this mixture environment.

\subsubsection{Probabilistic Problem Formulation}

An alternative formulation of the problem statement for unknown environments is a one-step procedure that essentially stays within (Bayesian) probability calculus. In this case actions are treated as random variables and our goal is to determine a distribution $P(a_t|\hat{a}_{<t}, o_{<t})$
that tells us how to act depending on past actions $\hat{a}_{<t}$ and past observations $o_{<t}$. The distribution $P(a_t|\hat{a}_{<t}, o_{<t})$ is the pendant to the observational distribution $P(o_t|\hat{a}_{\leq t}, o_{<t})$ used for prediction. The only caveat is that past probabilistic actions, unlike past observations, have to be marked as causal interventions---denoted by $\hat{\circ}$ in causal calculus \cite{Pearl1996}. Given the models~\eqref{eq:prediction-models}, both distributions can then be expressed as mixture distributions
\begin{eqnarray}
P(o_t|\hat{a}_{\leq t}, o_{<t}) &=& \sum_\theta B(o_t|\theta,a_{\leq t}, o_{<t}) B(\theta|\hat{a}_{\leq t}, o_{<t}) \nonumber \\
P(a_t|\hat{a}_{<t}, o_{<t}) &=& \sum_\theta B(a_t|\theta,a_{<t}, o_{<t}) B(\theta|\hat{a}_{<t}, o_{<t}) , \nonumber
\end{eqnarray}
where the posterior is given by
\[
 B(\theta|\hat{a}_{<t}, o_{<t}) = \frac{B(o_{t-1}|\theta,a_{<t-1}, o_{<t-1}) B(\theta|\hat{a}_{<t-1}, o_{<t-1})}{\sum_{\theta'} B(o_{t-1}|\theta',a_{<t-1}, o_{<t-1}) B(\theta'|\hat{a}_{<t-1}, o_{<t-1})} .
\]
Sampling from $P(a_t|\hat{a}_{<t}, o_{<t})$ is equivalent to sampling a random belief $\theta$ from the posterior $B(\theta|\hat{a}_{<t}, o_{<t})$ and then acting according to $B(a_t|\theta,a_{<t}, o_{<t})$. This corresponds to a generalized Thompson sampling procedure, where first a random belief is sampled and then the optimal policy with respect to this belief is executed.
Effectively, the posterior is also the only place where causal calculus plays a role, as can be seen by the absence of the likelihood $B(\hat{a}_{t-1}|\theta,a_{<t-1}, o_{<t-1})$, which is equal to one. Intuitively, the reason for this is that the agent can be surprised about his past observations and learn from them, but he cannot be surprised about his own actions chosen by himself in the past. Past actions do not provide any information about the environment. As will be explained in more detail below, causal calculus deals exactly with inference problems when some random variables are intervened or set by the inference machine itself.
Importantly, this result is obtained solely by applying basic probability and causal calculus.

\section{Policy Uncertainty}\label{sec:policy-uncertainty}

While subjective expected utility is formally appealing as a principle for the construction of adaptive agents, its strict application is in practice often problematic. This is mainly due to two reasons:
\begin{enumerate}
\item \emph{Computational complexity.}
    The computations required to find the optimal solution (for instance, the computational complexity of solving the Bellman optimality equations) are prohibitive in general and scale exponentially with the length of the horizon: the time complexity of the search algorithm is $\order(|\fs{A} \times \fs{O}|^T)$. The problem is tractable only in very special cases under assumptions that reduce the effective size of the problem.
\item \emph{Causal precedence of policy choice.}
    The choice of the policy has to be made before the interaction with the environment starts. That is, an agent has to have a unique optimal policy before it has even interacted once with the environment. An optimal policy constructed by the maximum SEU
    principle is therefore a very risky bet, as a lot of resources have to be spent
    before any evidence exists that the underlying model or prior is adequate.
\end{enumerate}

Because of these two reasons, it is practically often impossible or questionable to apply the maximum SEU principle. In the following, we investigate how to weaken the formal assumptions of the policy construction method.

\subsection{Policy search}

Given a problem specification in terms of the predictive model and the utility function, the task of a policy search method is to calculate a policy that approximates the optimal policy.

More specifically, let $\pi$ be a parameter in a set $\Pi$ indexing the set of candidate policies
\begin{equation}\label{eq:policy-models}
    B(a_t|\pi, a_{<t}, o_{<t})
\end{equation}
analogous to the prediction models \eqref{eq:prediction-models} indexed by $\theta \in \Theta$. \emph{Then, in the most general case, a policy search method returns a probability distribution~$B(\pi)$ over~$\Pi$ representing the uncertainty over the optimal policy parameters.} If the algorithm solves the maximum SEU problem, then the support of this distribution will exclusively cover the set of optimal policies $\Pi^\ast \subset \Pi$.
Otherwise there remains uncertainty over the optimal policy parameters.

Policies can also be parameterized in terms of the predictive model.
In particular, we will assume that for each $\theta \in \Theta$ there is a \emph{known}
optimal policy $\pi \in \Pi$, such that one can construct a function
$b:\Theta \rightarrow \Pi$ that maps each $\theta$ into some $\pi \in \Pi$. Uncertainty
over the environment can then directly be translated into policy uncertainty, such that
any point in the convex hull $H(\Theta)$ can be mapped to a corresponding point
in the convex hull $H(\Pi)$ spanned by the policies $\pi \in \Pi$.

\subsection{The Exploration-Exploitation Trade-Off}

Many policy search methods do not explicitly deal with the uncertainty over the policy parameters. Some methods only return a point estimate $\hat{\pi} \in \Pi$. It is obvious that the greedy usage of the estimate $\hat{\pi}$ leads to sub-optimal performance, since for all $\hat{\pi}$ that are not in the set $\Pi^\ast$ of optimal policies, one has that
\[
    \expect_B[U|\pi^\ast] > \expect_B[U|\hat{\pi}]
\]
where
\[
    \expect_B[U|\pi]
    = \sum_{a_{\leq T}, o_{\leq T}}
        B(a_{\leq T}, o_{\leq T}|\pi) U(a_{\leq T}, o_{\leq T})
\]
is the SEU with respect to the policy parameter $\pi \in \Pi$. For instance, reinforcement learning algorithms \citep{Sutton1998} start from a randomly initialized point estimate $\hat{\pi}_0$ of the optimal policy and then generate refined point estimates $\hat{\pi}_1, \hat{\pi}_2, \hat{\pi}_3, \ldots$ in each time step $t = 1, 2, 3, \ldots$ using the data provided by experience. In order to converge to the optimal policy, these algorithms have to deal with the exploration-exploitation trade-off. This means that the agents cannot just greedily act according to these point estimates; instead, they have to produce explorative actions as well, that is, actions that deviate from the current estimate of the optimal policy---for instance producing optimistic actions based on UCB \citep{Lai1985, Auer2002}.

Let $B_t(\pi)$ denote the posterior distribution over the optimal policy at time $t$. Then,
\[
    B_t(\pi) = B_t(b(\theta)) = B_t(b^{-1}(\pi)),
\]
where $b^{-1}(\pi) \subset \Theta$ is the pre-image of $\pi$. Hence, this shows that finding the optimal policy amounts to finding the  pre-image of $\pi^\ast$, such that the distribution over the policy space becomes the delta function
\[
    B_t(\pi) =
    \begin{cases}
        1 & \text{if $\theta^\ast \in b^{-1}(\pi)$,} \\
        0 & \text{else}
    \end{cases}
\]
where $\theta^\ast \in \Theta$ is the true prediction model defined in~\eqref{eq:true-prediction-model}. This highlights the essence of the exploration-exploitation trade-off: any action issued by the agent has to respect the uncertainty over the policy parameter---otherwise they are biased. In particular, if the agent acts greedily (i.e.\ it treats the estimate $\hat{\pi}$ as if it were the true policy parameter) then it is overfitting the experience; likewise, an agent having excessive uncertainty is underfitting. From a frequentist point of view, this reveals that the exploration-exploitation trade-off is nothing else but the \emph{bias-variance trade-off} \citep{Geman1992} in policy space. This suggests that just like Bayesian modeling naturally balances the bias-variance trade-off
by creating estimators that are probability distributions instead of point estimates, Bayesian modeling of the exploration-exploitation trade-off leads to a Bayes-causal solution for generalized Thompson sampling.

\subsection{Bayes-Causal Solution}\label{sec:bayes-causal-solution}

It is important to note that the concept of the bias is conditional on the true parameter---which is unknown when the designer is uncertain about the environment. This is not a problem from a Bayesian point of view, because the best estimator of the policy parameter is its posterior distribution \citep{Gelman2003}. Hence, instead of dealing with the exploration-exploitation dilemma by introducing explorative actions, one can directly use the posterior distribution over the policy parameter as an estimate.

To see how to do this, note that, by virtue of the mapping $\pi = b(\theta)$, the policies are independent of the policy parameter when the environment parameter is known:
\[
    B(a_t|b(\theta), a_{<t}, o_{<t})
    = B(a_t|\theta, a_{<t}, o_{<t}).
\]
Hence, each $\theta \in \Theta$ indexes a dynamical model given by the distributions over interaction sequences
\begin{align*}
    B(a_{\leq T}, o_{\leq T}|\theta)
    &= \prod_{t=1}^T B(a_t|\theta, a_{<t}, o_{<t})
        B(o_t|\theta, a_{\leq t}, o_{<t}).
\end{align*}
Given the dynamical models and their prior probabilities, the designer can form the Bayesian mixture model
\begin{equation}\label{eq:interaction-model}
    B(a_{\leq T}, o_{\leq T})
    = \sum_\theta B(\theta) B(a_{\leq T}, o_{\leq T}|\theta)
\end{equation}
where the sum spans all the parameters in $\Theta$. The mixture models in $H(\Theta) \setminus \Theta$ need not be considered here, since it is assumed that $\theta^\ast \in \Theta$.

\subsubsection*{Actions as Causal Interventions}

The designer can directly use the probabilistic model~\eqref{eq:interaction-model} to characterize an agent with policy uncertainty. There is a \emph{caveat} though when actions are treated as random variables. It is clear that the observations produced by the environment update the agent's state of knowledge about the environment. However, the actions are set by the agent itself and hence they do not provide information about the environment.

The theory that deals with the distinction between exogenous and endogenous information is \emph{statistical causality} \citep{Pearl2009, Spirtes2000}. Observations change the information state by regular Bayesian conditioning, whereas actions constitute causal interventions followed by Bayesian conditioning. To calculate the effect of an intervention, the causal model, i.e.\ the unique factorization of the joint distribution into conditional probabilities matching the causal dependencies over the random variables, is required to be known. In our setup, this is straightforward: first, the environment secretly chooses a true parameter $\theta^\ast \in \Theta$, and then the interactions $a_1, o_1, a_2, o_2, \ldots$ follow chronologically.

Formally, this means that the posterior probabilities over the environment parameters are given by
\[
    B(\theta|\hat{a}_{<t}, o_{<t})
\]
rather than the more familiar expression $B(\theta|a_{<t}, o_{<t})$, where the ``hat''-notation $\hat{a}_t$ denotes a causal intervention \citep{OrtegaBraun2010}.

For our needs, it is enough to consider the following simple method to calculate the effect of causal interventions:
\begin{enumerate}
\item[1.] Expand the probabilities in terms of the joint distribution.
\item[2.] Rewrite the joint distribution as the causal factorization.
\item[3.] Remove the intervention tags from the intervened random variables that are in the probability conditions.
\item[4.] Replace each conditional probability having an intervened variable in the argument by a delta function over its chosen value.
\end{enumerate}

Applying these four steps to the posterior probabilities over the environment parameters yields
\begin{small}
\begin{align}
    \nonumber
    &B(\theta|\hat{a}_{<t}, o_{<t}) \\
    \nonumber
    &\stackrel{(1.)}{=}
        \frac{ B(\theta, \hat{a}_{<t}, o_{<t}) }
             { \sum_{\theta'} B(\theta', \hat{a}_{<t}, o_{<t}) } \\
    \nonumber
    &\stackrel{(2.)}{=}
        \frac{ B(\theta) \prod_{k=1}^t
               B(\hat{a}_k|\theta, \hat{a}_{<k}, o_{<k})
               B(o_k|\theta, \hat{a}_{\leq k}, o_{<k}) }
             { \sum_{\theta'} B(\theta') \prod_{k=1}^t
               B(\hat{a}_k|\theta', \hat{a}_{<k}, o_{<k})
               B(o_k|\theta', \hat{a}_{\leq k}, o_{<k}) } \\
    \nonumber
    &\stackrel{(3.)}{=}
        \frac{ B(\theta) \prod_{k=1}^t
               B(\hat{a}_k|\theta, a_{<k}, o_{<k})
               B(o_k|\theta, a_{\leq k}, o_{<k}) }
             { \sum_{\theta'} B(\theta') \prod_{k=1}^t
               B(\hat{a}_k|\theta', a_{<k}, o_{<k})
               B(o_k|\theta', a_{\leq k}, o_{<k}) } \\
    \label{eq:intervened-posterior}
    &\stackrel{(4.)}{=}
        \frac{ B(\theta) \prod_{k=1}^t
               B(o_k|\theta, a_{\leq k}, o_{<k}) }
             { \sum_{\theta'} B(\theta') \prod_{k=1}^t
               B(o_k|\theta', a_{\leq k}, o_{<k}) }.
\end{align}
\end{small}
This equation shows that beliefs are updated only using the observations, and that actions are treated ``as if they were known beforehand'', thus providing no evidence.

Likewise, note that
\begin{equation}\label{eq:intervened-action}
    B(a_t|\theta, \hat{a}_{<t}, o_{<t})
    = B(a_t|\theta, a_{<t}, o_{<t}).
\end{equation}

Using~\eqref{eq:intervened-posterior} and~\eqref{eq:intervened-action}, we get the probability of issuing action $a_t \in \fs{A}$:
\begin{equation}\label{eq:bcr}
    B(a_t|\hat{a}_{<t}, o_{<t})
    = \sum_\theta B(a_t|\theta, a_{<t}, o_{<t}) B(\theta|\hat{a}_{<t}, o_{<t}).
\end{equation}
The important fact about~\eqref{eq:bcr} is it was derived only from probability theory and causal calculus by assuming policy uncertainty over a set of policies.
We can therefore define the policy and prediction models of the agent as
\begin{equation}\label{eq:adaptive-agent}
    \begin{aligned}
    P(a_t|a_{<t}, o_{<t}) &:= B(a_t|\hat{a}_{<t}, o_{<t}) \\
    P(o_t|a_{\leq t}, o_{<t}) &:= B(o_t|\hat{a}_{\leq t}, o_{<t}).
    \end{aligned}
\end{equation}

The construction of an adaptive agent with policy uncertainty then proceeds analogous to a Bayesian inference process.
\begin{enumerate}
\item First, we define a set of prediction models $B(o_t|\theta, a_{<t}, o_{<t})$
and policy models $B(a_t|\theta, a_{\leq t}, o_{<t})$, where each policy model is optimal for a particular environment $\theta$. In the case of inference
the latent random variable $\theta \in \Theta$ corresponds to the hypothesis.
\item Second, we choose some prior probabilities $B(\theta)$ to model our prior uncertainty.
\item Third, we use the distribution $B(a_t|\hat{a}_{<t}, o_{<t})$ as the agent's adaptive policy $P(a_t|a_{<t}, o_{<t})$, and the distribution $B(o_t|\hat{a}_{\leq t}, o_{<t})$ as the agent's adaptive predictor $P(o_t|a_{\leq t}, o_{<t})$.
\end{enumerate}
Thus, Thompson sampling is used in every time step to sample an action $a_t$ from the predictive distribution $B(a_t|\hat{a}_{<t}, o_{<t})$.

\section{Convergence \& Co-Adaptation}

In \cite{OrtegaBraun2010e}, the limit behavior of a Thompson sampling agent \eqref{eq:adaptive-agent} was investigated. Assuming that there exists a belief  $P(o_t|\theta,a_{\leq t}, o_{<t})$ that perfectly models the environment $Q_\theta$ such that $P(o_t|\theta,a_{\leq t}, o_{<t})=Q_\theta(o_t|a_{\leq t}, o_{<t})$, then the agent \eqref{eq:adaptive-agent} converges in the sense that $P(a_t|a_{<t},o_{<t}) \rightarrow P(a_t|\theta,a_{<t},o_{<t})$ almost surely as $t \rightarrow \infty$ if the interaction system $(P,Q)$ fulfills certain ergodicity requirements and all policies $P(a_t|\theta,a_{<t},o_{<t})$ are consistent. Roughly speaking, the first requirement ensures that the agent can recover from any initial mistakes, and the second requirement ensures that all predictors $B(o_t|\theta,a_{\leq t},o_{<t})$ that make the same predictions for the tail of the observation sequence are coupled to the same policy $B(a_t|\theta,a_{<t},o_{<t})$. Thus, the same beliefs imply the same behaviors.

But what happens if the environment is also adaptive? As long as the agent has a model $B(o_t|\theta,a_{\leq t},o_{<t})$ that captures the adaptive behavior of the environment nothing fundamentally changes. However, the agent might not have a model about the \emph{adaptive} behavior of the environment, while still having a pretty good idea about the environment's preferences. This is typically the case in game theory \cite{Gibbons1992,Osborne1999}, where the agent knows the other agent's best response function, but has no model of the other agent's adaptive behavior. In the simplest one-shot simultaneous move games the best response functions are given by
\begin{eqnarray}
BR_1[Q(o)] &=& \arg \max_{P(a)} \sum_{a,o} P(a) Q(o) U(a,o) \nonumber \\
BR_2[P(a)] &=& \arg \max_{Q(o)} \sum_{a,o} P(a) Q(o) V(a,o) \nonumber ,
\end{eqnarray}
where player $1$'s best response $BR_1$ is a distribution $P(a)$ over actions $a$ that depends on agent $2$'s probability $Q(o)$ of emitting $o$, and where $U$ and $V$ are the payoff functions for player $1$ and $2$ respectively. A Nash equilibrium $\left( P^*(a), Q^*(o) \right)$ is a fix point of these coupled equations \cite{Nash1950}, where each individual player has no incentive to change his distribution, that is
\begin{eqnarray}
\sum_{a,o} P^*(a) Q^*(o) U(a,o) &\geq& \sum_{a,o} P(a) Q^*(o) U(a,o) \quad \forall P(a) \nonumber \\
\sum_{a,o} P^*(a) Q^*(o) V(a,o) &\geq& \sum_{a,o} P^*(a) Q(o) V(a,o) \quad \forall Q(o) \nonumber .
\end{eqnarray}
How such equilibria are reached is not subject in classic equilibrium game theory. But, if such games are repeated over and over again, evolutionary game theory suggests that these equilibria appear as fix points of adaptation dynamics like the replicator equations \cite{Weibull1995}. As Bayesian inference can also be viewed as some kind of replicator dynamics \cite{Shahlizi2009,Harper2009}, this provides an interesting starting point to study the emergence of Nash equilibria when two adaptive Thompson sampling agents interact.

When both agents are adaptive according to \eqref{eq:adaptive-agent}, we can decompose the policies $P(a)$ and $Q(o)$ into mixture distributions, such that
\begin{eqnarray}
P(a)&=&\sum_\theta P(a|\theta) P(\theta) \nonumber \\
Q(o)&=&\sum_\xi Q(o|\xi) Q(\xi) \nonumber ,
\end{eqnarray}
where $P(a|\theta) = BR_1[Q(o)]$ and $Q(o|\xi) = BR_2[P(a)]$ and $P(\theta)$ and $Q(\xi)$ are the prior distributions over the possible behaviors.
Moreover, both agents will have predictive models over the other agent's behavior, such that
\begin{eqnarray}
P(o)&=&\sum_\theta P(o|\theta) P(\theta) \nonumber \\
Q(a)&=&\sum_\xi Q(a|\xi) Q(\xi) \nonumber ,
\end{eqnarray}
where $P(o|\theta)=Q(o|\xi)$ and $Q(a|\xi)=P(a|\theta)$. In the following we will assume that there exists at least one pair ($\theta^*, \xi^*)$ where the model $P(o|\theta^*)$ perfectly matches $Q(o|\xi^*)$ such that $P(o|\theta^*)=Q(o|\xi^*)$ and at the same time the model $Q(a|\xi^*)$ perfectly matches $P(a|\theta^*)$ such that $P(a|\theta^*)=Q(a|\xi^*)$. In this case both agents can predict the other agent's behavior, which means that there will be no drive to change the posteriors $P(\theta|D)$ and $Q(\xi|D)$,  given some past experience $D$.
Then, both agents should ``lock in'' when their posteriors are sufficiently close to $\delta_{\theta,\theta^*}$ and $\delta_{\xi,\xi^*}$. Formally, we define a pair $(\xi^*, \theta^*)$ to be a strict Nash equilibrium if
\begin{eqnarray}
\sum_o Q(o|\xi^*) \log \frac{Q(o|\xi^*)}{P(o|\theta^*)} &<& \sum_o Q(o|\xi^*) \log \frac{Q(o|\xi^*)}{P(o)} \qquad \forall P(o) \nonumber \\
\sum_a P(a|\theta^*) \log \frac{P(a|\theta^*)}{Q(a|\xi^*)} &<& \sum_o P(a|\theta^*) \log \frac{P(a|\theta^*)}{Q(a)} \qquad \forall Q(a) \nonumber .
\end{eqnarray}
This corresponds to a lock-in of the predictive models that drive the adaptation process.

In order to study convergence to $(\theta^*,\xi^*)$,
we determine the difference in relative entropy between the predictive distribution $P(o)$ at time $t$ and the generative distribution $Q(o|\xi^*)$ and the relative entropy between the predictive distribution $P(o|D)$ at time $t+1$ and the generative distribution $Q(o|\xi^*)$---after observing $D$ at time step $t$. This difference can serve as a Lyapunov function to show convergence, if we require that
\[
\Delta KL = \sum_D Q(D|\xi^*) D_{KL}\left(Q(o|\xi^*) \| P(o|D) \right) - D_{KL}\left(Q(o|\xi^*) \| P(o) \right) < 0 .
\]
Given the predictive distribution $P(o)=\sum_\theta P(o|\theta) P(\theta)$ at time $t$ and the predictive distribution $P(o|D)=\sum_\theta P(o|\theta) P(\theta|D)$ at time $t+1$, we get
\begin{eqnarray}
\Delta KL &=& \sum_D Q(D|\xi^*) \sum_o Q(o|\xi^*) \left[ \log \frac{Q(o|\xi^*)}{\sum_\theta P(o|\theta) P(\theta|D)} - \log \frac{Q(o|\xi^*)}{\sum_\theta P(o|\theta) P(\theta)} \right] \nonumber \\
&=& \sum_D Q(D|\xi^*) \sum_o Q(o|\xi^*) \log \frac{\sum_\theta P(o|\theta) P(\theta)}{\sum_\theta P(o|\theta) P(\theta|D)} \nonumber \\
&\leq& \sum_D Q(D|\xi^*) \sum_o Q(o|\xi^*) \left[ \log \frac{\sum_\theta P(o|\theta) P(\theta)}{P(o|\theta^*) P(\theta^*|D)} \right] \nonumber \\
&=& \sum_o Q(o|\xi^*) \log \frac{\sum_\theta P(o|\theta) P(\theta)}{P(o|\theta^*)} - \sum_D Q(D|\xi^*) \log P(\theta^*|D) \nonumber \\
&=& \sum_o Q(o|\xi^*) \log \frac{\sum_\theta P(o|\theta) P(\theta)}{P(o|\theta^*)} - \sum_D Q(D|\xi^*) \log \frac{P(D|\theta^*)P(\theta^*)}{\sum_{\theta'} P(D|\theta')P(\theta')} \nonumber \\
&=& \underbrace{\sum_o Q(o|\xi^*) \log \frac{\sum_\theta P(o|\theta) P(\theta)}{P(o|\theta^*)}}_{<0 \textrm{   (Nash property)}} \underbrace{+ \sum_D Q(D|\xi^*) \log \frac{\sum_{\theta'} P(D|\theta')P(\theta')}{P(D|\theta^*)}}_{<0 \textrm{   (Nash property)}} \underbrace{- \log P(\theta^*)}_{\geq 0} \nonumber
\end{eqnarray}
If the prior weight $P(\theta^*)$ is close to one then the last positive term $-\log P(\theta^*)$ is close to zero and the two other terms will dominate, making the whole expression negative, thus, implying convergence. The argument can also be extended to the case where $D$ is generated by $\sum_\xi Q(D|\xi)Q(\xi)$ instead of $Q(D|\xi^*)$, depending again on the weight of $Q(\xi^*)$ compared to the other weights $Q(\xi)$, i.e. how close the other agent is to the Nash policy. The same argument is then repeated for player $2$ who uses the prediction model $Q(a|\xi)$ to model $P(a|\theta)$.

\begin{figure}[tb]
\begin{center}
    \label{fig:matchingpennies}
    \includegraphics[scale=0.4]{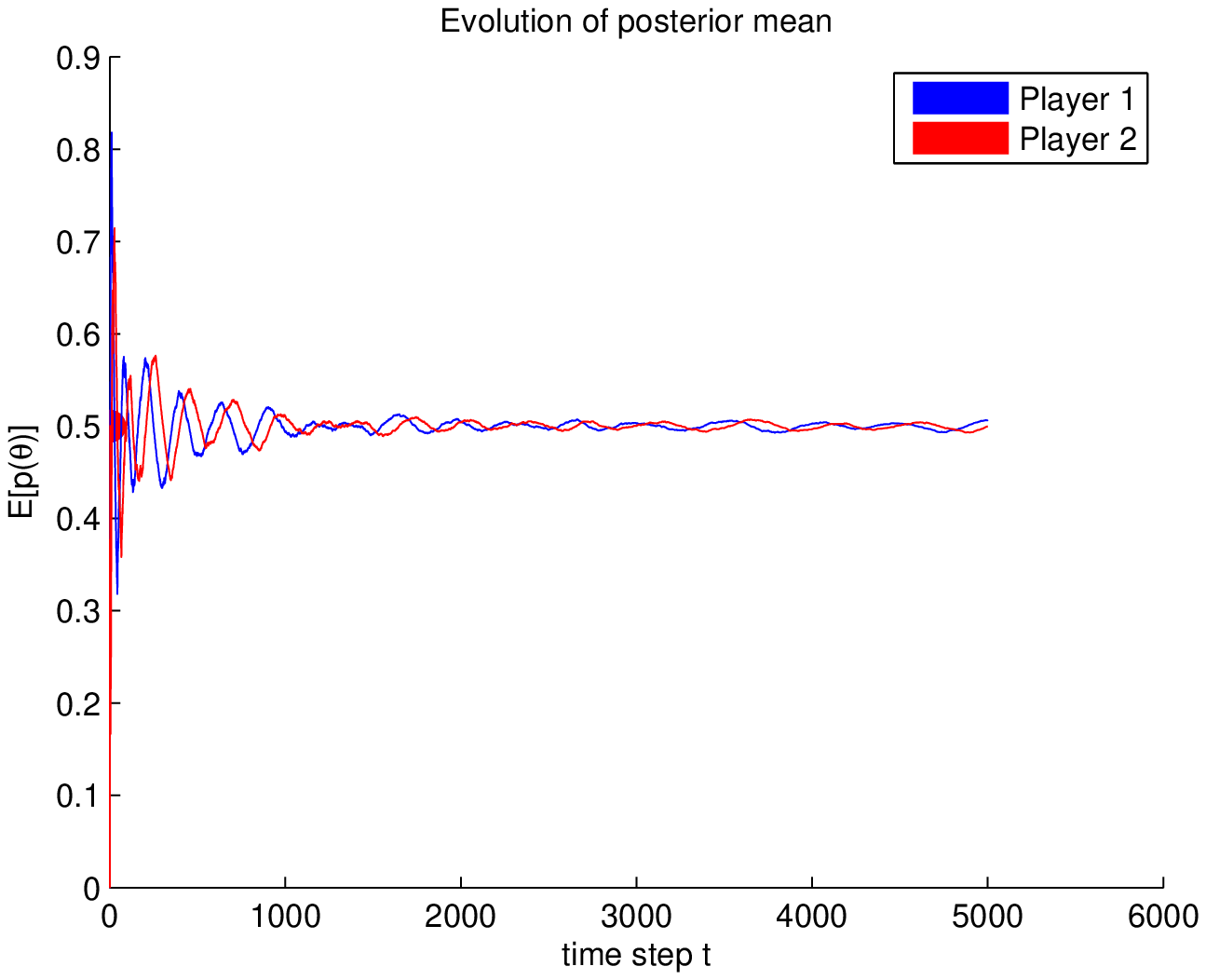}
    \includegraphics[scale=0.4]{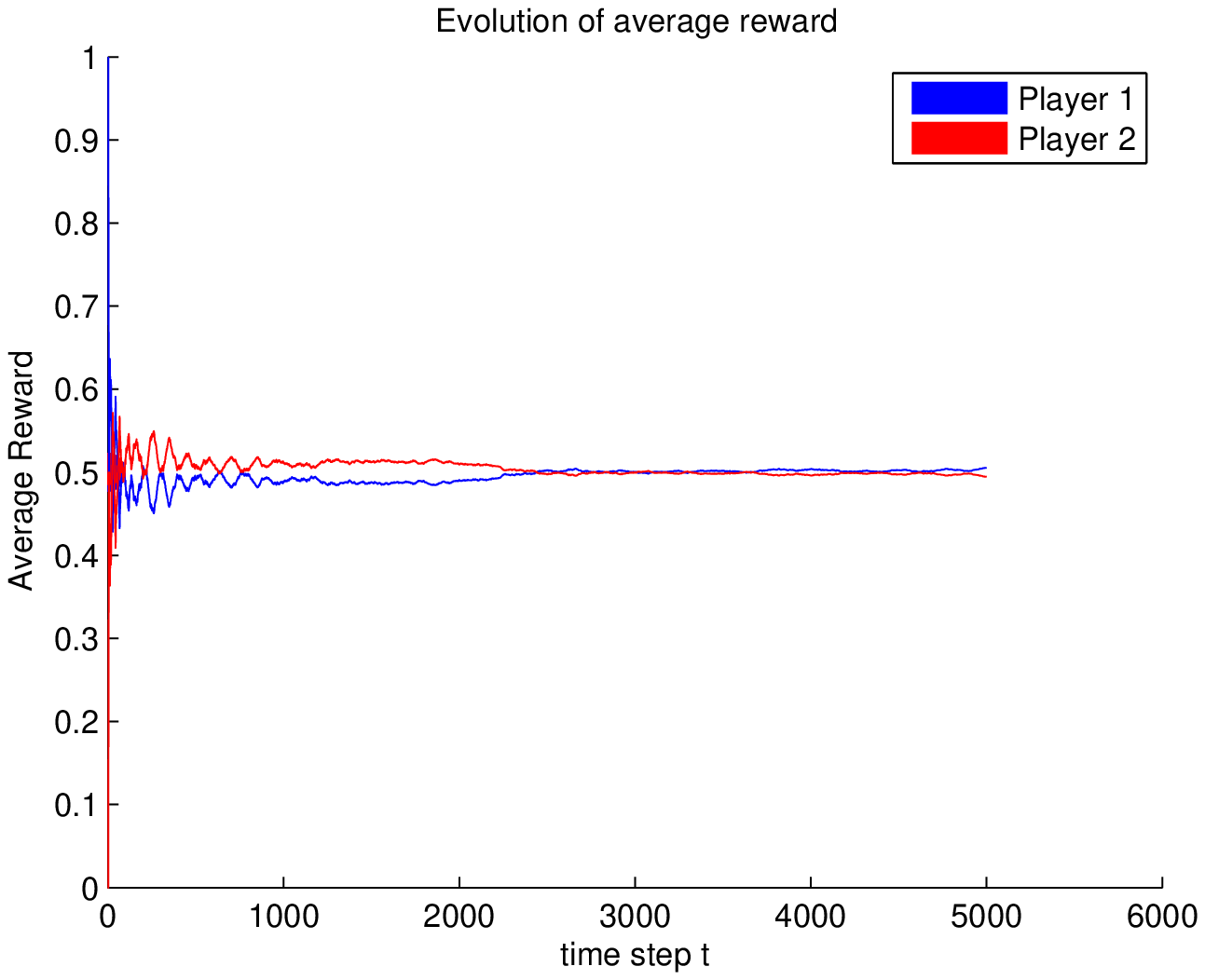}
    \caption{Matching pennies game with two Thompson sampling agents. The left figure shows the evolution of the mean of the posterior distributions over $\theta$ and $\xi$ for a single run. The right figure shows the average mean reward obtained by each player.}
    \label{fig:ambiguity}
\end{center}
\end{figure}

Consider as an example the matching pennies game \cite{Osborne1999}, where each player has a penny and must decide whether to secretely turn their penny to heads or tails. Player $1$ wins if both pennies show heads or both show tails, whereas player $2$ wins if the pennies are unmatched. The payoff matrices of the game are
\begin{displaymath}
\left(
\begin{array}{cc}
(+1, -1) & (-1,+1) \\
(-1, +1) & (+1,-1)
\end{array}
\right) ,
\end{displaymath}
where the first number in each cell is the payoff for player $1$ and the second number is the payoff for player $2$, and the columns of the matrix correspond to the choice of heads or tails of player $1$, and the rows of the matrix correspond to the choice of heads or tails of player $2$. The best responses are then
\begin{displaymath}
P(a=H|\theta) = \left\{ \begin{array}{ll}
0 & \textrm{if $\theta<1/2$}\\
\frac{1}{2} & \textrm{if $\theta=1/2$}\\
1 & \textrm{if $\theta>1/2$}
\end{array} \right\}
\end{displaymath}
and
\begin{displaymath}
Q(o=H|\xi) = \left\{ \begin{array}{ll}
1 & \textrm{if $\xi<1/2$}\\
\frac{1}{2} & \textrm{if $\xi=1/2$}\\
0 & \textrm{if $\xi>1/2$}
\end{array} \right\}
\end{displaymath}
and the predictive models are $P(o|\theta)=\theta$ and $Q(a|\xi)=\xi$. For the priors we assume the uniform distributions $P(\theta)=1$ and $Q(\xi)=1$, such that the posteriors are
\[
P(\theta; a_1,a_2) = \frac{\theta^{a_1-1} (1-\theta)^{a_2-1}}{B(a_1,a_2)}
\]
and
\[
Q(\xi; b_1,b_2) = \frac{\xi^{b_1-1} (1-\xi)^{b_2-1}}{B(b_1,b_2)},
\]
where $B(\cdot,\cdot)$ is the beta function, and $a_1$ and $a_2$ is the number of heads and tails played by player $2$, whereas $b_1$ and $b_2$ is the number of heads and tails played by player $1$.
In this case the only $(\theta^*,\xi^*)$ pair that is a Nash equilibrium is in the $50:50$ case, because only then do action and prediction model fit for both agents. In each time step the agents sample $\theta$ and $\xi$ respectively from their posteriors $P(\theta|\cdot)$ and $Q(\xi|\cdot)$ and act with their best response to this sample. In Figure~\ref{fig:matchingpennies} it can be seen how they co-adapt and converge to the Nash equilibrium.

\section{Causal induction}

Agent~\eqref{eq:adaptive-agent}  can be thought of as a probabilistic superposition of models $\theta$, where each model $\theta$ is characterized by a likelihood model $P(o_t|\theta,\g{{a}o}_{<t}{a}_t)$ and a policy model $P(a_t|\theta,\g{{a}o}_{<t})$.
In previous applications we assumed that all models $\theta$ have the same causal structure,
i.e. considering multivariate random variables $a_t$ and $o_t$, we assumed that the same variables $a_t$ are intervened for all $\theta$ and the same causal model is used to predict the consequences of these interventions on the observational variables $o_t$.
However, this need not be the case. In principle, different models $\theta$ could represent different causal structures and suggest intervention of different variables. Such a setup can be used for causal induction.

Imagine, for example, we are given a device with two light bulbs, one green ($X$) and one red ($Y$), whose states obey a hidden mechanism that correlates them positively. Moreover, the device has switches that allow us controlling the state of either bulb.. We encode the ``on'' and ``off'' states of the green light as $X=x$ and $X=\neg x$ respectively. Analogously, $Y=y$ and $Y=\neg y$ denote the ``on'' and ``off'' states of the red light. We are interested in the explanatory power of two competing hypotheses: either ``green causes red'' ($\Theta=\theta$) or ``red causes green'' ($\Theta=\neg \theta$).

One of the main methods to deal with problems of causal inference is the framework of causal graphical models \cite{Pearl2009}. Given a graph that represents a causal structure, we can intervene this graph and ask questions about the probabilities of the variables in the graph. However, in causal induction we would like to discover the causal structure itself, that is we would like to do inference over a multitude of graphs representing different causal structures \cite{Heckerman97abayesian}. If one would like to represent the problem of causal discovery graphically, the main challenge is that the model $\Theta$ is a random variable that controls the causal structure itself. That is, a tentative graphical representation would be
\begin{center}
    \small
    \psfrag{x1}[c]{$X$}
    \psfrag{x2}[c]{$Y$}
    \psfrag{x3}[c]{$Y$}
    \psfrag{x4}[c]{$X$}
    \psfrag{x5}[c]{$\Theta$}
    \psfrag{a1}[c]{meta-level}
    \includegraphics{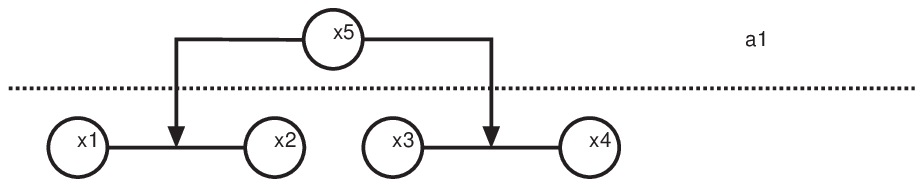}
\end{center}
which cannot be analyzed using the mathematical framework of graphical models alone because the random variable $\Theta$ operates on a meta-level of the graphical model over $X$ and $Y$. In fact, different causal structures have to be investigated by different graphical models, that is the inference process over different causal structures cannot be represented in one and the same graphical model. However, this difficulty can be overcome by using a probability tree to model the causal structure over the random events \cite{OrtegaNips2011}. Probability trees can encode alternative causal realizations, and in particular alternative causal hypotheses \cite{Shafer1996}. All random variables are then of the same type---no distinctions between meta-levels are needed.

\subsubsection*{Representation}

\begin{figure}[tbp]
\begin{center}
    \scriptsize
    \psfrag{l1}[l]{a)}
    \psfrag{l2}[l]{b)}
    \psfrag{p00}[c]{$\frac{1}{2}$}
    \psfrag{p01}[c]{$\frac{1}{2}$}
    \psfrag{p02}[c]{$\frac{1}{2}$}
    \psfrag{p03}[c]{$\frac{1}{2}$}
    \psfrag{p04}[c]{$\frac{1}{2}$}
    \psfrag{p05}[c]{$\frac{1}{2}$}
    \psfrag{p06}[c]{$\frac{3}{4}$}
    \psfrag{p07}[c]{$\frac{1}{4}$}
    \psfrag{p08}[c]{$\frac{1}{4}$}
    \psfrag{p09}[c]{$\frac{3}{4}$}
    \psfrag{p10}[c]{$\frac{3}{4}$}
    \psfrag{p22}[c]{$\frac{1}{4}$}
    \psfrag{p12}[c]{$\frac{1}{4}$}
    \psfrag{p13}[c]{$\frac{3}{4}$}
    \psfrag{p14}[c]{$\frac{3}{16}$}
    \psfrag{p15}[c]{$\frac{1}{16}$}
    \psfrag{p16}[c]{$\frac{1}{16}$}
    \psfrag{p17}[c]{$\frac{3}{16}$}
    \psfrag{p18}[c]{$\frac{3}{16}$}
    \psfrag{p19}[c]{$\frac{1}{16}$}
    \psfrag{p20}[c]{$\frac{1}{16}$}
    \psfrag{p21}[c]{$\frac{3}{16}$}
    \psfrag{a00}[c]{$\Omega$}
    \psfrag{a01}[c]{$\theta$}
    \psfrag{a02}[c]{$\neg \theta$}
    \psfrag{a03}[c]{$x$}
    \psfrag{a04}[c]{$\neg x$}
    \psfrag{a05}[c]{$y$}
    \psfrag{a06}[c]{$\neg y$}
    \psfrag{a07}[c]{$y$}
    \psfrag{a08}[c]{$\neg y$}
    \psfrag{a09}[c]{$y$}
    \psfrag{a10}[c]{$\neg y$}
    \psfrag{aaa}[c]{$x$}
    \psfrag{a12}[c]{$\neg x$}
    \psfrag{a13}[c]{$x$}
    \psfrag{a14}[c]{$\neg x$}
    \psfrag{q00}[c]{$\frac{1}{2}$}
    \psfrag{q01}[c]{$\frac{1}{2}$}
    \psfrag{q02}[c]{$\bf{1}$}
    \psfrag{q03}[c]{$\bf{0}$}
    \psfrag{q04}[c]{$\frac{1}{2}$}
    \psfrag{q05}[c]{$\frac{1}{2}$}
    \psfrag{q06}[c]{$\frac{3}{4}$}
    \psfrag{q07}[c]{$\frac{1}{4}$}
    \psfrag{q08}[c]{$\frac{1}{4}$}
    \psfrag{q09}[c]{$\frac{3}{4}$}
    \psfrag{q10}[c]{$\bf{1}$}
    \psfrag{q22}[c]{$\bf{0}$}
    \psfrag{q12}[c]{$\bf{1}$}
    \psfrag{q13}[c]{$\bf{0}$}
    \psfrag{q14}[c]{$\frac{3}{8}$}
    \psfrag{q15}[c]{$\frac{1}{8}$}
    \psfrag{q16}[c]{$0$}
    \psfrag{q17}[c]{$0$}
    \psfrag{q18}[c]{$\frac{1}{4}$}
    \psfrag{q19}[c]{$0$}
    \psfrag{q20}[c]{$\frac{1}{4}$}
    \psfrag{q21}[c]{$0$}
    \psfrag{b00}[c]{$\Omega$}
    \psfrag{b01}[c]{$\theta$}
    \psfrag{b02}[c]{$\neg \theta$}
    \psfrag{b03}[c]{$x$}
    \psfrag{b04}[c]{$\neg x$}
    \psfrag{b05}[c]{$y$}
    \psfrag{b06}[c]{$\neg y$}
    \psfrag{b07}[c]{$y$}
    \psfrag{b08}[c]{$\neg y$}
    \psfrag{b09}[c]{$y$}
    \psfrag{b10}[c]{$\neg y$}
    \psfrag{baa}[c]{$x$}
    \psfrag{b12}[c]{$\neg x$}
    \psfrag{b13}[c]{$x$}
    \psfrag{b14}[c]{$\neg x$}
    \includegraphics[scale=0.8]{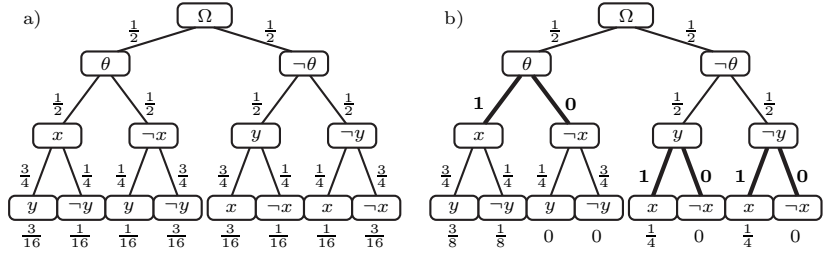}
    \caption{a) An exemplary probability tree to represent the agent's prediction model about its environment. The probability of a realization (written under the leaves) is calculated by multiplying the probabilities starting from the root until a leave is reached. Note that the two hypotheses are statistically indistinguishable. b) The probability tree resulting from (a) after setting $X=x$.}
    \label{fig:tree}
\end{center}
\end{figure}

We can use probability trees to represent the prediction model that the agent has about its environment.
An exemplary probability tree for our problem is depicted in Figure~\ref{fig:tree}. In this tree, each (internal) node is interpreted as a causal mechanism; hence a path from the root node to one of the leaves corresponds to a particular sequential realization of causal mechanisms. The logic underlying the structure of this tree is as follows:
\begin{enumerate}
    \item \emph{Causal precedence:} A node causally precedes its descendants. For instance, the root node corresponding to the sure event $\Omega$ causally precedes all other nodes.
    \item \emph{Resolution of variables:} Each node resolves the value of a random variable. For instance, given the node corresponding to $\Theta=\theta$ and $X=\neg x$, either $Y=y$ will happen with probability $\prob(y|\theta,\neg x) = \frac{1}{4}$ or $Y=\neg y$ with probability $\prob(\neg y|\theta,\neg x) = \frac{3}{4}$.
    \item \emph{Heterogeneous order:} The resolution order of random variables can vary across different branches. For instance, $X$ precedes $Y$ under $\Theta=\theta$, but $Y$ precedes $X$ under $\Theta=\neg \theta$. This allows modeling different causal hypotheses.
\end{enumerate}
While the probability tree represents our subjective model explaining the order in which the random values are resolved, it does not necessarily correspond to the temporal order in which the events are revealed to us. So for instance, under hypothesis $\Theta=\theta$, the value of the variable $Y$ might be revealed before $X$, even though $X$ causally precedes $Y$; and the hypothesis $\Theta$, which precedes both $X$ and $Y$, is never observed.

\subsubsection*{Interventions}

The importance of interventions to detect causal structure is illustrated in Figure~\ref{fig:tree}, as the observational probabilities are completely symmetric for the two halves of the tree.
Suppose we observe that both lights are on. Have we learned anything about their causal dependency? A brief calculation shows that this is not the case because the posterior probabilities are equal to the prior probabilities:
\small
\begin{eqnarray}
    \prob(\theta|x,y)
        &=& \frac{\prob(y|\theta,x)\prob(x|\theta)\prob(\theta)}
                {\prob(y|\theta,x)\prob(x|\theta)\prob(\theta)
                + \prob(x|\neg \theta,y)\prob(y|\neg \theta)\prob(\neg \theta)} \nonumber \\
        &=& \frac{\frac{3}{4} \cdot \frac{1}{2} \cdot \frac{1}{2}}
               {\frac{3}{4} \cdot \frac{1}{2} \cdot \frac{1}{2}
               + \frac{3}{4} \cdot \frac{1}{2} \cdot \frac{1}{2}}
        = \frac{1}{2} = \prob(\theta) \nonumber.
\end{eqnarray}
\normalsize
This makes sense intuitively, because by just observing that the two lights are on, it is statistically impossible to tell which one caused the other. The only way to extract causal information is then to intervene, paraphrased as ``no causes in, no causes out'' \citep{Cartwright1994} or ``to find out what happens when you kick the system, you have to kick the system'' \citep{Box1966}.
Thus, we now repeat our experiment, but this time \emph{we} turn on the green light ($X=x$). We reflect this choice by changing all the mechanisms that resolve the random variable $X$, placing all the probability mass on the outcome $X=x$ (see Figure~\ref{fig:tree}b). Assume that we subsequently observe that the second light is on. Then, the posterior probabilities are
\small
\begin{eqnarray}
    \prob(\theta|\hat{x},y)
        &=& \frac{\prob(y|\theta,\hat{x})\prob(\hat{x}|\theta)\prob(\theta)}
                {\prob(y|\theta,\hat{x})\prob(\hat{x}|\theta)\prob(\theta)
                + \prob(\hat{x}|\neg \theta,y)\prob(y|\neg \theta)\prob(\neg \theta)} \nonumber \\
        &=& \frac{\frac{3}{4} \cdot 1 \cdot \frac{1}{2}}
               {\frac{3}{4} \cdot 1 \cdot \frac{1}{2}
                + 1 \cdot \frac{1}{2} \cdot \frac{1}{2}}
        = \frac{3}{5} \nonumber,
\end{eqnarray}
\normalsize
where $\hat{x}$ is Pearl's notation to indicate a causal intervention of $X$. Since $P(\theta) < P(\theta|\hat{x},y)$, we have gathered evidence favoring the hypothesis ``green causes red''. This was only possible because our intervention introduced a statistical asymmetry among the two hypotheses that did not exist before.

\subsubsection*{Thompson Sampling}

\begin{figure}[tb]
\begin{center}
    \includegraphics[scale=0.4]{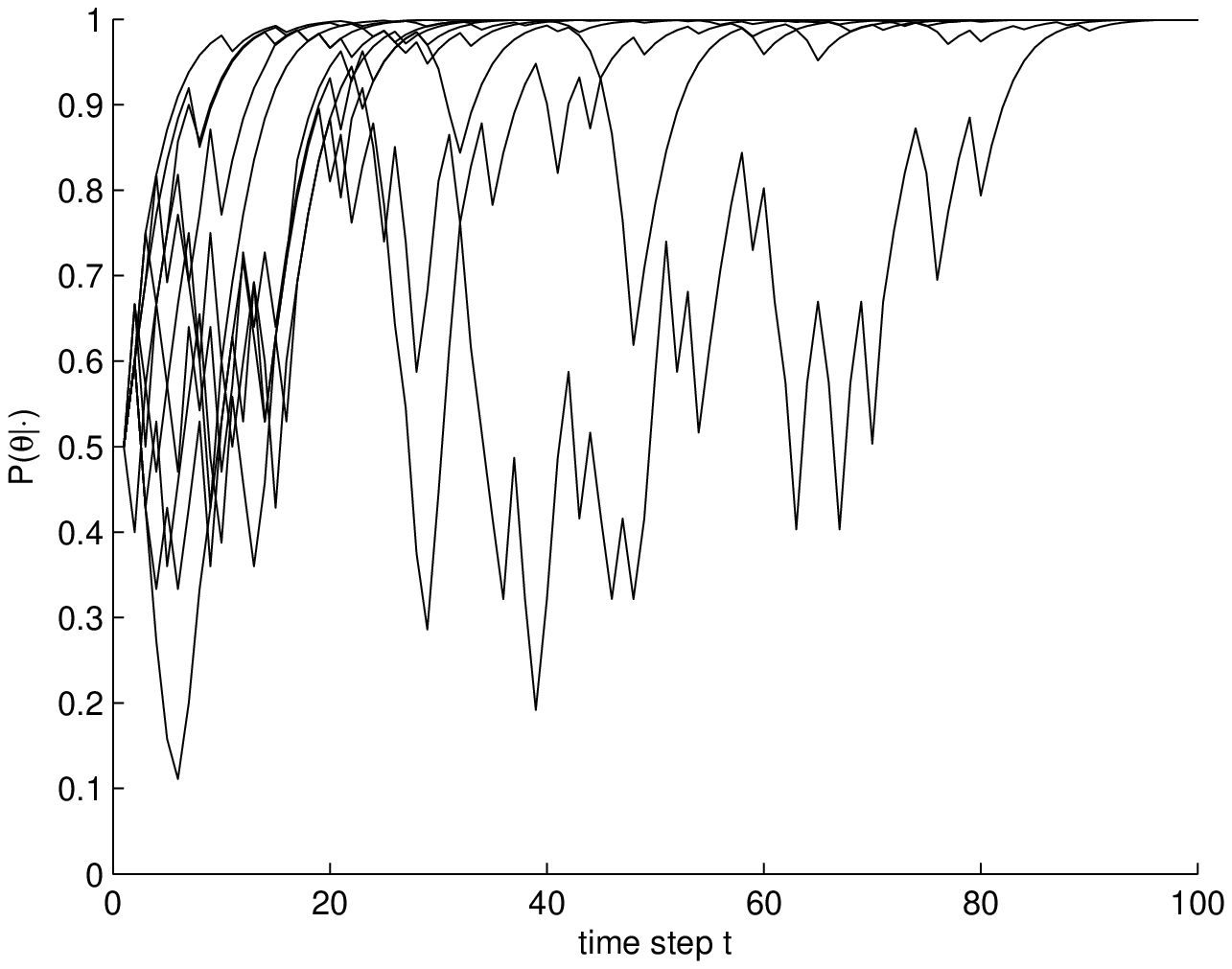}
    \includegraphics[scale=0.4]{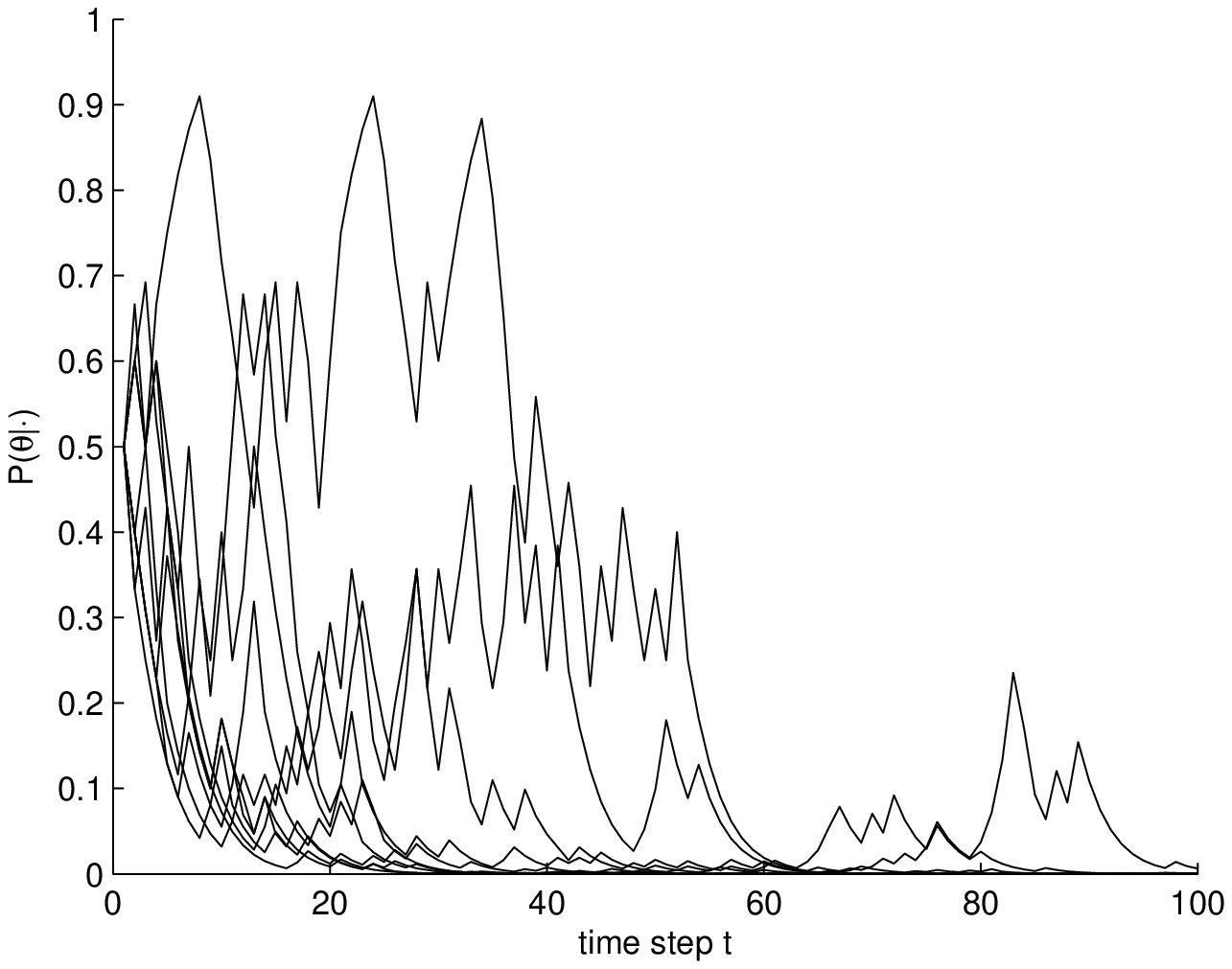}
    \caption{Thompson sampling for causal induction. (Left) Posterior distribution $P(\theta|\cdot)$ for $10$ runs     when the true system is given by $\Theta=\theta$. (Right) Posterior distribution $P(\theta|\cdot)$ for $10$ runs when the true system is given by $\Theta=\neg \theta$. In both cases the agent is able to identify the causal structure of the environment with high confidence when $P(\theta|\cdots)$ is close to one or zero respectively.}
    \label{fig:causality}
\end{center}
\end{figure}

Naturally, multiple interventions and observations can be executed in consecution.
In this case Thompson sampling is used in each time step to decide which policy model to use, which implies the decision which variables to intervene.
Then, after the intervention, all variables are revealed simultaneously at every time step of the inference process. The update of the observational probabilities is done the same way as in the one step case, taking into account which variables were intervened. A simulation of the repeated Thompson sampling process for causal induction of our example system is shown in Figure~\ref{fig:causality}.

\section{Discussion}\label{sec:discussion}

\label{sec:related-approaches}

Equations~\eqref{eq:adaptive-agent} was first derived in \cite{OrtegaBraun2010e} as the optimal solution to the adaptive coding problem given actions and observations as \emph{Bayesian rule for control}. In practice, it is implemented by sampling an environment parameter $\hat{\theta}_t$ for each time step from the posterior distribution $B(\theta|\hat{a}_{<t}, o_{<t})$, and then treating it as if it was the true parameter---that is, issuing the action $a_t$ from $B(a_t|\hat{\theta}_t, a_{<t}, o_{<t})$. This action-sampling method where beliefs are randomly instantiated was first proposed as a heuristic in \citep{Thompson1933} and is now known as \emph{Thompson sampling}.
Equations~\eqref{eq:adaptive-agent} therefore provides a method for generalized Thompson sampling applicable to adaptive sequential decision-making problems.

The main contribution of this paper is to examine three features of such generalized Thompson sampling. First, we provide an argument showing that Thompson sampling is a natural consequence of a Bayesian treatment of policy uncertainty. Policy uncertainty arises whenever an agent is trying to find an optimal policy, but is unable to do so---for example due to computational constraints---, even though the agent might have a coarse idea about the optimum, which can be expressed as a probability distribution. The Bayesian treatment of this uncertainty is analogous to Bayesian estimation in the case of pure observation problems. The Bayes-optimal estimator in this case is not point estimate, but a distribution, which forgoes the bias-variance dilemma. Similarly, in the case of actions, the exploration-exploitation trade-off can be circumvented by Thompson sampling from probabilistic policies expressed as Bayesian mixture distributions.

Second, we investigated co-adaptation of two adaptive Thompson sampling agents. We could demonstrate that such agents converge to Nash equilibria, if the parameterized policy set they are choosing from is given by the parameterized best response functions. This approach also generalizes previous models of fictitious play \cite{Brown1951,Fudenberg1998} that best-respond to the observed frequency of the opponent's play rather than best-responding to their randomized beliefs about the opponent.  Therefore, adaptive Thompson sampling agents might provide a useful modeling tool for evolutionary game theory \cite{Weibull1995} and learning in games in general \cite{Fudenberg1993}.

Third, we could demonstrate that generalized Thompson sampling can also be applied to the problem of causal induction, by designing policy and prediction models with different causal structures. 
This way generalized Thompson sampling can be used as a general method for causal induction that is Bayesian in nature. It is based on the idea of combining probability trees \citep{Shafer1996} with interventions \citep{Pearl2009} for predicting the behavior of a manipulated system with multiple causal hypotheses. Both the interventions and the constraints on the causal hypotheses introduce statistical asymmetries that permit the extraction of causal information. Unlike frameworks that aim to extract causal information from observational data alone \cite{Shimizu2006,Griffiths2009,Janzing2010}, the proposed method is designed for agents that interact with their environment and use these interactions to discover causal relationships.

So far Thompson sampling has been mainly applied to multi-armed bandit problems.
Multi-armed bandits can be represented by a parameter $\theta$ that summarizes the statistical properties of the reward obtained for each lever. Reward distributions range from Bernoulli to Gaussian (with unknown mean and variance), and they can also depend on the particular context or state \cite{Graepel2010, May2011, Granmo2010, Scott2010}. In particular, the work of \cite{May2011} and the work of \cite{Granmo2010} prove asymptotic convergence of Thompson sampling. The work of \cite{Chapelle2011} presents empirical results that show Thompson sampling is highly competitive, matching or outperforming popular methods such as UCB \cite{Lai1985,Auer2002}.

Another class of problems, where Thompson sampling has been applied in the past, are Markov decision processes (MDPs). MDPs can be described by parameterizing the dynamics and reward distribution (model-based) \cite{Strens2000} or by directly parameterizing the \emph{Q-table} (model-free) \cite{Dearden1998, OrtegaBraun2010}. The first approach samples a full description of an MDP, solves it for the optimal policy, and then issues the optimal action. This is repeated in each time step. The second approach avoids the  computational overhead of solving for the optimal policy in each time step by directly doing inference on the Q-tables. Actions are chosen by picking the one having the highest Q-value for the current state. The same ideas can also be applied to solve adaptive control problems with linear system equations, quadratic cost functions and Gaussian noise \cite{OrtegaBraun2010c}.

\subsection{Optimality}

One of the main arguments is that the derivation presented in Section~\ref{sec:bayes-causal-solution} shows that generalized Thompson sampling is not just a heuristic method, but that it can be derived under the assumption of policy uncertainty---simply by applying probability theory and causal calculus. This Thompson sampling approach differs from the formulation of adaptive control problems as control problems with known environments that require the maximization of a subjective expected utility criterion---compare Section~\ref{sec:SEUformulation}.
The difference between the two approaches can be highlighted by contrasting the two one-step decision scenarios depicted in Figure~\ref{fig:ambiguity}. The goal is to predict the outcome of a biases coin with payoffs $\$1$ and $\$0$ for a right and wrong guess respectively. A rational decision maker places bets (shown inside speech bubbles) such that his subjective expected utility is maximized. These subjective beliefs are delimited within dotted boxes.
\begin{figure}[tb]
\begin{center}
    \small
    \psfrag{la}[c]{a)}
    \psfrag{lb}[c]{b)}
    \psfrag{l1}[c]{H}
    \psfrag{l2}[c]{T}
    \psfrag{t1}[c]{$\tfrac{1}{4}$}
    \psfrag{t2}[c]{$\tfrac{3}{4}$}
    \psfrag{t3}[c]{$\tfrac{3}{4}$}
    \psfrag{t4}[c]{$\tfrac{1}{4}$}
    \psfrag{b1}[c]{T}
    \psfrag{b2}[c]{H}
    \psfrag{b3}[c]{H}
    \includegraphics{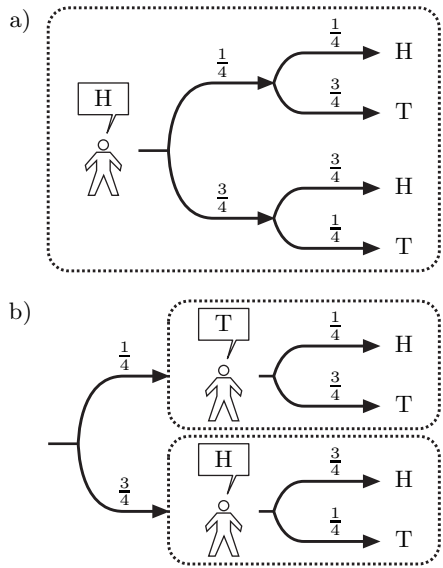}
    \caption{Comparison between (a) a fully rational decision maker and (b) a decision maker with random beliefs.}
    \label{fig:ambiguity}
\end{center}
\end{figure}

The difference between the two becomes clear by inspecting the expected utility in each case: they are
\begin{align*}
    &\max_{P'} \sum_\theta P(\theta)
        \Bigl\{ \sum_o P'(a|\theta) P(o|\theta, a) U(o) \Bigr\},
    &&\text{(a)}
    \\ \text{and} \quad
    &\sum_\theta P(\theta) \max_{P'}
        \Bigl\{ \sum_o P'(a|\theta) P(o|\theta, a) U(o) \Bigr\}
    &&\text{(b)}
\end{align*}
respectively. Here it is clearly seen that the difference between the two lies in the order in which we apply the expectation (over the environment parameter) and the maximization operator. Both cases can be explained in terms of optimality. However, in (a), decision-maker picks his action taking into account the uncertainty over the bias, while in (b), the decision-maker picks his action only after his beliefs over the coin bias are instantiated---that is, \emph{he is optimal w.r.t. his random beliefs.}

The difference between probabilities that one takes into account when making a decision versus the probabilities that are not (i.e.\ they are \emph{immeasurable}) has been first proposed by \cite{Knight1921}. The classical decision theories of \cite{Neumann1944} and \cite{Savage1954} only consider known probabilities that are reasoned about inside the max-operation. Another example where random beliefs play a crucial role is in \emph{games with incomplete information} \cite{Osborne1999}. Here, having incomplete information about the other player leads to a infinite hierarchy of meta-reasoning about the other player's strategy. To avoid this difficulty, Harsanyi introduced \emph{Bayesian games} \cite{Harsanyi1967}. In a Bayesian game, incomplete knowledge is modeled by randomly instantiating the player's types, after which they choose their strategies optimally---thus eliminating the need for recurrent reasoning about the other players' strategy.

Maintaining and updating Bayesian probabilities is an optimally efficient way to deal with uncertainty---be it with respect to the policy or the environment \citep{OrtegaBraun2010}.
Therefore, the central claim is that having random beliefs---as formalized by generalized Thompson sampling---can be considered optimal \emph{under the constraint} of having policy uncertainty---uncertainty that is inevitable whenever we are unable to compute the optimal policy.
Having policy uncertainty effectively weakens the two assumptions of the maximum expected utility principle: the optimal policy can be chosen and refined during interactions, and the computational complexity is lower.

The operational distinction of having policy uncertainty has important algorithmic consequences. When there is policy uncertainty, the belief of the decision-maker is itself a random variable. This means that the very policy is undefined until the random variable is resolved. Hence, the computation of the optimal policy can be delayed and determined dynamically. It is precisely this fact that is (implicitly) exploited in popular reinforcement learning algorithms, and explicitly in the algorithms based on random beliefs. This is in stark contrast to the case when there is no policy uncertainty, where the policy is pre-computed and static.

\subsubsection*{Adaptive Coding and the Kullback-Leibler Divergence}
\label{subsubsec:KL}

Even though the maximization is inside the expectation in case of random belief approaches to decision-making, there is another outer maximization or optimality criterion implicit, analogous to the case of Bayesian inference that is known to optimize Kullback-Leibler divergences. Therefore, it is useful to think about the adaptive control problem as an adaptive coding and inference problem \cite{OrtegaBraun2010e,OrtegaNips2012}. In terms of the initial problem statement in Section~\ref{sec:problem}, the question then is: How can the designer construct a
 system $P$ defined by $P(o_t|a_{\leq t}, o_{<t})$ and $P(a_t|a_{< t}, o_{<t})$ such that its behavior is as close as possible to
the custom-made system $B(o_t|\theta,a_{\leq t}, o_{<t})$ and $B(a_t|\theta,a_{< t}, o_{<t})$ under any realization of $Q_\theta$? Using the Kullback-Leibler divergence as a distance measure, we can formulate a variational problem in $P$
\begin{equation}\label{eq:minimization-causal}
    P \define \arg \min_{\prob} \Bigg\{ \limsup_{t \rightarrow \infty}
        \sum_{\theta} P(\theta) \sum_{\tau=1}^t \bigl(
            D_m^{a_\tau}(\prob) + D_m^{o_\tau}(\prob) \bigr) \Bigg\}
\end{equation}
with
\begin{align*}
    D_\theta^{a_t}(\prob) &= \sum_{a_{<t},o_{<t}}
        B(a_{<t},o_{<t}|\theta) \sum_{a_t} B(a_t|\theta,a_{<t},o_{<t})
        \log \frac{ B(a_t|\theta,a_{<t},o_{<t}) }{ \prob(a_t|a_{<t},o_{<t}) } \\
    D_\theta^{o_t}(\prob) &= \sum_{a_{\leq t},o_{<t}}
        B(a_{\leq t},o_{<t}|\theta) \sum_{o_t} B(o_t|\theta,a_{\leq t},o_{<t})
        \log \frac{ B(o_t|\theta,a_{\leq t},o_{<t}) }{ \prob(o_t|a_{\leq t},o_{<t})
        }.
\end{align*}
In the case of observations, this is a well-known variational principle for Bayesian inference, as it describes a predictor that requires, on average, the least amount of extra bits to capture informational surprise stemming from the behavior of the environment. In the case of actions, the same principle can be harnessed to describe resourceful generation of actions in a way that requires random bits with minimum length on average, when trying to match the optimal policy most suitable for the unknown environment \cite{Mackay2003}.

\subsection{Evolutionary game theory}

When dealing with adaptive agents, one of the most intriguing questions is what happens if two adaptive agents are coupled. Classic game theory does not really allow to address that question, as it abstracts away from learning and adaptation processes and focuses on fix point conditions for equilibria. In contrast, evolutionary game theory focuses on the dynamics that can lead to equilibria \cite{Weibull1995}. One of the most widely studied dynamics in evolutionary game theory are the so-called replicator equations
\[
x^i_{t+1} = \frac{x^i_t f_i(x)}{\sum_j x^j_t f_j(x)} ,
\]
where $x^i_t$ represents the proportion of type $i$ in a population  of individuals at time $t$. The vector $x_t=(x^1_t, \ldots, x^n_t)$ represents the population distribution, such that $\sum_j x_t^j = 1$. The function $f_i(x)$ denotes the fitness of type $i$, which depends on the population $x$. The proportion of individuals of type $i$ at the next time point $t+1$ is determined by the fitness share this type achieves compared to the population total.

Interestingly, there is a formal correspondence between the replicator dynamics and Bayesian inference \cite{Shahlizi2009,Harper2009}
\[
p(h|d) = \frac{p(h) p(d|h)}{\sum_{h'} p(h') p(d|h')},
\]
where $p(h)$ and $p(h|d)$ represents the prior and posterior probability mass allotted to hypothesis $h$.
The likelihood function $p(d|h)$ works as a fitness landscape. The posterior probability is determined by the likelihood fitness achieved compared to the overall evidence $P(d)=\sum_{h'} p(h') p(d|h')$.

In evolutionary game theory the fix points of the replicator dynamics have been studied extensively.
In particular, evolutionarily stable strategies \cite{smith82} have been shown to be a refinement of the common Nash equilibrium, in the sense that such Nash equilibria are stable with respect to perturbing mutant strategies.
Since generalized Thompson sampling as described in \eqref{eq:adaptive-agent} shares its form with Bayesian inference, the connection to evolutionary replicator equations is immediate. Therefore, we could apply very similar stability arguments in the case of two interacting adaptive agents \eqref{eq:adaptive-agent}, as previously applied in the case of the replicator dynamics. Generalized Thompson sampling might therefore also provide a useful tool in the future to study convergence of co-adaptation processes within the context of evolutionary game theory.

\subsection{Causality}

To construct the Bayes-causal solution in Section~\ref{sec:bayes-causal-solution}, we needed to treat actions as interventions. This raises the question about why this distinction was not made for deriving classical SEU solutions. More formally, under what conditions do the equations in~\eqref{eq:adaptive-agent} reduce to
\[
    \begin{aligned}
    P(a_t|a_{<t}, o_{<t}) &= B(a_t|a_{<t}, o_{<t}) \\
    P(o_t|a_{\leq t}, o_{<t}) &= B(o_t|a_{\leq t}, o_{<t}),
    \end{aligned}
\]
that is, with no interventions?

Since,
\begin{align*}
    B(a_t|a_{<t}, o_{<t})
    &= \sum_\theta B(a_t|\theta, a_{<t}, o_{<t})
        B(\theta|\hat{a}_{<t}, o_{<t}) \\
    B(o_t|a_{\leq t}, o_{<t})
    &= \sum_\theta B(o_t|\theta, a_{\leq t}, o_{<t})
        B(\theta|\hat{a}_{\leq t}, o_{<t}),
\end{align*}
determining the conditions boils down to analyzing when the equalities
\begin{eqnarray}
    B(\theta|a_{<t}, o_{<t}) &=& B(\theta|\hat{a}_{<t}, o_{<t}) \nonumber \\
    B(\theta|\hat{a}_{\leq t}, o_{<t}) &=& B(\theta|a_{\leq t}, o_{<t}) \nonumber
\end{eqnarray}
hold. Replacing both sides yields,
\begin{gather}
    \nonumber
    \frac{ B(\theta) \prod_{k=1}^t
           B(a_k|\theta, a_{<k}, o_{<k})
           B(o_k|\theta, a_{\leq k}, o_{<k}) }
         { \sum_{\theta'} B(\theta') \prod_{k=1}^t
           B(a_k|\theta', a_{<k}, o_{<k})
           B(o_k|\theta', a_{\leq k}, o_{<k}) } \\
    \label{eq:causal-condition}
    = \frac{ B(\theta) \prod_{k=1}^t
              B(o_k|\theta, a_{\leq k}, o_{<k}) }
            { \sum_{\theta'} B(\theta') \prod_{k=1}^t
              B(o_k|\theta', a_{\leq k}, o_{<k}) }
\end{gather}
Inspecting~\eqref{eq:causal-condition} we conclude that
\[
    B(a_k|\theta, a_{<k}, o_{<k})
    = \delta_{\bar{a}_k}(a_k),
\]
i.e.\ the actions have to be issued deterministically (but possibly history-dependent) from a unique policy. Intuitively speaking, this is because the operations of intervening and conditioning coincide when the random variables are deterministic.

\subsection{Open Problems}

There are important cases where random belief approaches can fail.
Indeed, it is easy to devise experiments where having policy uncertainty converges exponentially slower (or does not converge at all) than knowing the optimal policy.

Consider the following simple example: Environment~1 is a $k$-state MDP in which only $k$ consecutive actions $A$ reach a state with reward $+1$. Any interception with a $B$-action leads back to the initial state. A second environment which is like the first but actions $A$ and $B$ are interchanged. The optimal policy figures out the true environment in $k$ actions (either $k$ consecutive $A$'s or $B$'s). Consider now an agent with random beliefs: The optimal action in environment~1 is $A$, in environment~2 is $B$. A uniform ($\frac{1}{2},\frac{1}{2}$) prior over the two possible environments stays a uniform posterior as long as no reward has been observed. Hence, an agent with random beliefs chooses at each time-step $A$ and $B$ with equal probability. With this policy it takes about $2^k$ actions to accidentally choose a row of $A$'s (or $B$'s) of length~$k$. From then on the agent acts optimally too. Thus, the optimal policy converges in time $k$, while the agent with policy uncertainty needs exponentially longer. A simple way to remedy this problem is, of course, to sample random beliefs only every $k$ time steps. But this problem can be exacerbated in non-stationary environments. Take for instance, an increasing MDP with $k = \bigl\lceil 10\sqrt{t} \,\bigr\rceil$, in which the optimal policy converges in $100$ steps, while an agent with policy uncertainty would not converge at all in most realizations.

Although \cite{OrtegaBraun2010e} prove asymptotic convergence for general environments fulfilling a restrictive form of ergodicity condition, this condition needs to be weakened for the convergence proof to be applicable to most real problems. But it is clear that a form of ergodicity is required for an agent with policy uncertainty to be able to learn to act optimally. Intuitively, this means that an agent can only learn if the environment has temporally stable statistical properties.
Finally, determining the speed of convergence and the \emph{regret} is currently an open problem.

\section{Conclusion}

In this paper we have argued that policy uncertainty is a natural phenomenon that arises whenever there are not enough computational resources to apply the maximum SEU principle. We have shown that treating this uncertainty in a Bayesian way with actions as random variables that obey causal calculus naturally leads to Thompson sampling and its Bayesian generalization. This generalized Thompson sampling can be straightforwardly applied to evolutionary game theory and to the problem of causal induction.
As these random-belief approaches can be derived simply from probability theory and causal calculus we suggest that they should not be considered as mere heuristics but as well-founded principled approaches.


\section{Acknowledgments}
This study was supported by the DFG, Emmy Noether grant BR4164/1-1.


\bibliographystyle{vancouver}
\bibliography{bibliography}

\end{document}